\documentclass[11pt]{article}
\usepackage{graphicx}
\usepackage{bm}
\usepackage{dcolumn}
\usepackage[margin=0.5in]{geometry}

\usepackage{color}
\usepackage{a4wide}
\usepackage{authblk}
\usepackage{amsfonts}
\usepackage{amssymb}
\usepackage{amstext}
\usepackage{amsmath}
\usepackage{datetime}
\usepackage{pdfpages}
\usepackage[font={small}]{caption}
\usepackage{url}
\def\@{\partial}
\def\<{\langle}
\def\>{\rangle}

\newcommand{\ignore}[1]{\index{ignore}}
\usepackage[sort&compress,sectionbib]{natbib}
\bibpunct{(}{)}{,}{a}{}{,}
\def\cite#1{\citep{#1}}

\begin{document}
\DeclareGraphicsExtensions{.pdf}
 

\title {Distributed Recurrent Neural Forward Models
with Synaptic Adaptation for Complex Behaviors of Walking Robots}
\author[1,2, 4]{\small Sakyasingha Dasgupta \thanks{Correspondence: sakyasingha.dasgupta@riken.jp \underline{Current address}: Riken Brain Science Institute, 2-1 Hirosawa, Wako, Saitama, Japan}}
\author[2]{\small Dennis Goldschmidt}
\author[1,2]{\small Florentin W\"{o}rg\"{o}tter}
\author[2,3]{\small Poramate Manoonpong}
\affil[1]{\small Institute for Physics - Biophysics, George-Agust-University, G\"{o}ttingen, Germany}
\affil[2]{\small Bernstein Center for Computational Neuroscience, George-Agust-University, G\"{o}ttingen, Germany}
\affil[3]{\small Maersk Mc-Kinney Moller Institute, University of Southern Denmark, Odense, Denmark}
\affil[4]{\small Riken Brain Science Institute, 2-1 Hirosawa, Wako, Saitama, Japan}
\newdate{date}{02}{04}{2015}
\date{\displaydate{date}}

\maketitle
\begin{abstract}

Walking animals, like stick insects, cockroaches or ants, demonstrate a fascinating range of locomotive abilities and complex behaviors. The locomotive behaviors can consist of a variety of walking patterns along with adaptation that allow the animals to deal with changes in environmental conditions, like uneven terrains, gaps, obstacles etc. Biological study has revealed that such complex behaviors are a result of a combination of biomechanics and neural mechanism thus representing the true nature of embodied interactions. While the biomechanics helps maintain flexibility and sustain a variety of movements, the neural mechanisms generate movements while making appropriate predictions crucial for achieving adaptation. Such predictions or planning ahead can be achieved by way of internal models that are grounded in the overall behavior of the animal. Inspired by these findings, we present here, an artificial bio-inspired walking system which effectively combines biomechanics (in terms of the body and leg structures) with the underlying neural mechanisms. The neural mechanisms consist of 1) central pattern generator based control for generating basic rhythmic patterns and coordinated movements, 2) distributed (at each leg) recurrent neural network based adaptive forward models with efference copies as internal models for sensory predictions and instantaneous state estimations, and 3) searching and elevation control for adapting the movement of an individual leg to deal with different environmental conditions. Using simulations we show that this bio-inspired approach with adaptive internal models allows the walking robot to perform complex locomotive behaviors as observed in insects, including walking on undulated terrains, crossing large gaps as well as climbing over high obstacles. Furthermore we demonstrate that the newly developed recurrent network based approach to online forward models outperforms the adaptive neuron forward models, which have hitherto been the state of the art, to model a subset of similar walking behaviors in walking robots.

\end{abstract}

\section{Introduction} 

Walking animals show diverse locomotor skills to deal with a wide range of terrains and environments. These involve intricate motor control mechanisms with internal prediction systems and learning \citep{huston2011studying}, allowing them to effectively cross gaps \citep{blaesing2004stick}, climb over obstacles \citep{watson2002control}, and even walk on uneven terrain \citep{pearson1984characteristics}, \citep{cruse1976control}. These capabilities are realized by a combination of biomechanics of their body and neural mechanisms. The main components of these neural mechanisms include central pattern generators (CPGs), internal forward models, and limb-reflex control systems. The CPGs generate basic rhythmic motor patterns for locomotion, while the reflex control employs direct sensory feedback \citep{pearson1984characteristics}. However, it is argued that biological systems need to be able to predict the sensory consequences of their actions in order to be capable of rapid, robust, and adaptive behavior. As a result, similar to the observations in vertebrate brains \citep{kawato1999internal}, insects can also employ internal forward models as a mechanism to predict their future state (predictive feedbacks) given the current state or sensory context (sensory feedback) and the control signals (efference copies), in order to shape the motor patterns for adaptation \citep{webb2004neural},\citep{mischiati2015internal}. Essentially, such a forward model acts as an internal feedback loop, that uses a copy of the motor command, in order to predict the expected sensory input. Comparing this to the actual input, appropriate modulations of this signal or adaptive behaviors can be carried out.

In order to make such accurate predictions of future actions to satisfy changing environmental demands, the internal forward models require some degree of memory of the previous sensory-motor information. However, given that, such motor control happens on a very fast timescale, keeping track of temporal information is integral to such very short-term memory processes. Reservoir-based recurrent neural networks (RNNs) \citep{maass2002real}, \citep{jaeger2004harnessing}, \citep{sussillo2009generating},  with their inherent ability to deal with temporal information and fading memory of sensory stimuli, thus provide a suitable platform to model such internal predictive mechanisms. Taking this perspective, here, we utilize a newly developed model of self-adaptive reservoir networks (SARN) \citep{dasgupta2013information}, \citep{dasgupta2015thesis}, to act as the forward models for sensorimotor prediction. This works in conjunction with other neural mechanisms for motor control and generates complex adaptive locomotion in an artificial walking robotic system. Specifically, by exploiting the adaptive recurrent layer of our model it is possible to achieve complex motor transformations at different walking gaits, which is significantly difficult to achieve by currently existing adaptive forward models employed with walking robots \citep{manoonpong2013neural}, \citep{dearden2005learning}, \citep{schroder2010using}. 

We present for the first time a distributed forward model architecture using six SARN-based forward models on a hexapod robot, each of which is for sensory prediction and state estimation of an individual robot leg. The outputs of the models are compared with foot contact sensory signals (actual sensory feedback) and  the differences between them are used for motor adaptation, in an online manner. This is integrated as part of the neural mechanism framework consisting of  1) single central pattern generator-based control for generating basic rhythmic patterns and coordinated movements, 2) distributed reservoir forward models and 3) searching and elevation action control for adapting the movement of an individual leg based on the forward model predictions, in order to deal with changing environmental conditions. 

In the following section we describe the architectural setup of the neural mechanisms used for the design of adaptive locomotion control in a walking robot, along with a description of the simulated hexapod robot AMOS II and the modular robot control environment used as the development platform  for our proposed control system. In section 3, we present the materials and methods used in this study. Specifically, we introduce the setup and implementation of the distributed reservoir-based adaptive forward model, with details of the learning procedure. Section 4 presents experimental results of the learning mechanism and the resulting behaviors of the simulated hexapod AMOS II on different complex locomotion scenarios likes crossing a large gap, walking on uneven (rough) terrains, and overcoming obstacles. The results obtained from the reservoir based forward models are juxtaposed with the previous state of the art adaptive neuron forward models setup. Finally, in section 5, we discuss our results and provide an outlook of further future directions.

\section{Neural Mechanisms for Complex Locomotion} 

\begin{figure}
\begin{center} 
\includegraphics[width = \linewidth]{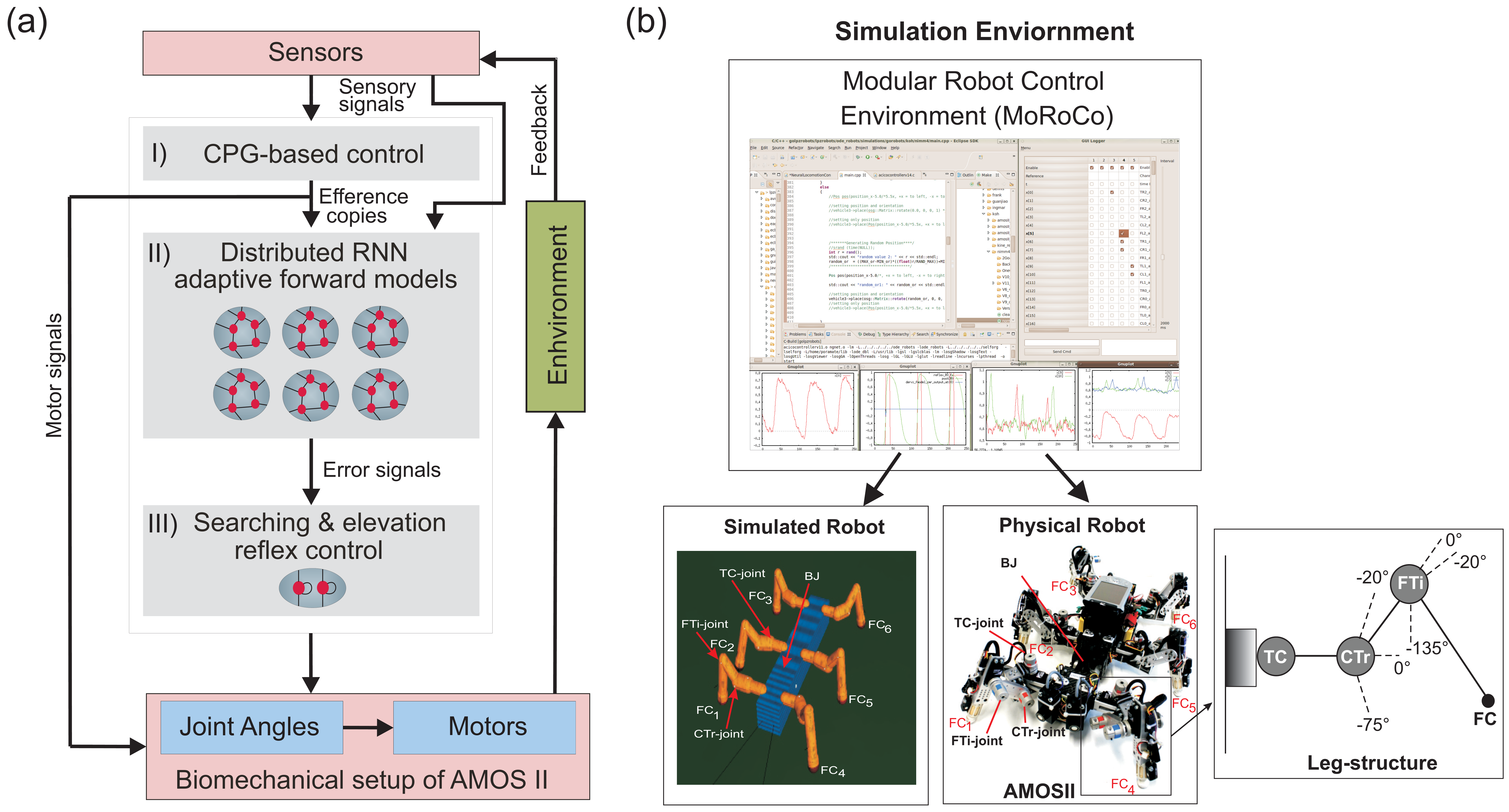} \caption{(a) The closed-loop architectural diagram of an artificial bio-inspired walking system consisting of the sensors (i.e., proprioceptive and exteroceptive sensors) that receive environmental inputs and feedback, the neural mechanisms (i,ii,iii) for adaptive locmotion control, and the biomechanical setup of the hexapod robot AMOSII (i.e., six 3-jointed legs, a segmented body structure with one active backbone joint (BJ), actuators, and passive compliant components \citep{manoonpong2013neural}). (b) Modular Robot Control Environment embedded in the LPZRobots simulation toolkit \citep{der2012lpzrobots}, \citep{hesse2012modular}. The simulation environment provides the main testbed for developing the controller, testing it on the simulated hexapod robot, and finally transferring it to the physical agent. Here we evaluate our model and results primarily on the simulated robot (bottom left), which accurately embodies the characteristics of its physical equivalent, AMOS II robot (bottom right). Here, $FC_1$, $FC_2$, $FC_3$, $FC_4$, $FC_5$, and $FC_6$ are foot contact sensors installed in the robot legs, which are used as the main sensory stimuli compared against the predicted signal from the RNN-based (reservoir) forward models. Each leg (right inset) consists of three joints: the innermost thoraco-coxal (TC-) joint enables forward and backward movements, the middle coxa-trochanteral (CTr-) joint enables elevation and depression of the leg, and the outermost femur-tibia (FTi-) joint enables extension and flexion of the tibia. The morphology of these multi-jointed legs were designed based on a cockroach leg \citep{zill2004load}. The front and back parts of the body are connected with a backbone joint (BJ) which primarily allows upwards and downwards tilting of the front body segment. Thus this is used for climbing and gap crossing purposes. More details on BJ control for climbing can be found in \citep{goldschmidt2014biologically}. }\label{fig:neuralmechanisms}
\end{center}
\end{figure}

The neural mechanisms (Figure~\ref{fig:neuralmechanisms} a) for locomotion control, are designed based on a modular architecture, such that, they comprise of, i) central pattern generator (CPG)-based control, ii) reservoir-based adaptive forward models, and iii) searching and elevation action control. The CPG-based control and the searching and elevation control have been previously discussed in detail in \citep{manoonpong2013neural}, thus here we will only provide a brief overview of these mechanisms, while the reservoir-based adaptive forward models, which forms the main topic of this work, will be presented in detail in the following section.

The CPG-based control primarily generates a variety of rhythmic patterns and coordinates all leg joints of a simulated hexapod robot AMOSII (Figure~\ref{fig:neuralmechanisms} (b)), thereby, leading to a multitude of different behavioral patterns and insect-like leg movements. The patterns include omnidirectional walking and insect-like gaits \citep{manoonpong2013neural}. All these patterns can be set manually, or autonomously driven by exteroceptive sensors, like a camera \citep{zenker2013visual}, a laser scanner \citep{kesper2013obstacle}, or range sensors. While the CPG-based control provides versatile autonomous behaviors, the searching and elevation control at each leg uses the accumulated error signals provided by the reservoir-based adaptive forward models in order to adapt the movement of an individual leg of the robot and deal with changes in environmental conditions. 

The CPG-based control (see supplementary Figure 1 for detailed description) itself is designed as a modular neural network that consists mainly of the following four elements:
\begin{enumerate}
\item CPG mechanism with neuromodulation for generating different rhythmic signals. Inspired by biological findings, here the CPG circuit is designed as a two-neuron fully connected recurrent network \citep{pasemann2003so}, such that using different external neuromodulatory inputs different walking gaits can be achieved. 
\item CPG post-processing units (PCPG) for shaping CPG output signals.
\item Phase switching network (PSN) and velocity regulating networks (VRNs) for walking directional control. 
\item Motor neurons with embedded fixed delay lines for transmitting motor commands to all leg joints of AMOS II. These delay lines are utilized to realize the inter-limb coordination, in which they introduce phase differences between the transmitted signals to all leg joints. As a result, the desired walking gait can be achieved. 
\end{enumerate}

The searching and elevation control at each leg, consist of single recurrent neurons that receive the difference (instantaneous error) between the predicted forward model signal and the actual sensory feedback. Due to the recurrent self-connection, this error is accumulated over time. The accumulated error can then be used to either extend specific leg joints in order to get better foothold (searching action) during the stance phase, or elevate further to overcome obstacles during the swing phase (see Figure~\ref{fig:testing_FM} (e) in section~\ref{sec:learn_fm}). All neurons in the CPG-based control and the searching and elevation control are modeled as discrete-time rate-coded neurons with tan-hyperbolic and piece-wise linear activation functions (see \citep{manoonpong2013neural} for details), respectively. They were updated with a frequency of~$\approx$~27~Hz.


\section{Materials \& Method}

\subsection{Reservoir-based Distributed Adaptive Forward Models}

\begin{figure}
\begin{center}
\includegraphics[scale = 0.20]{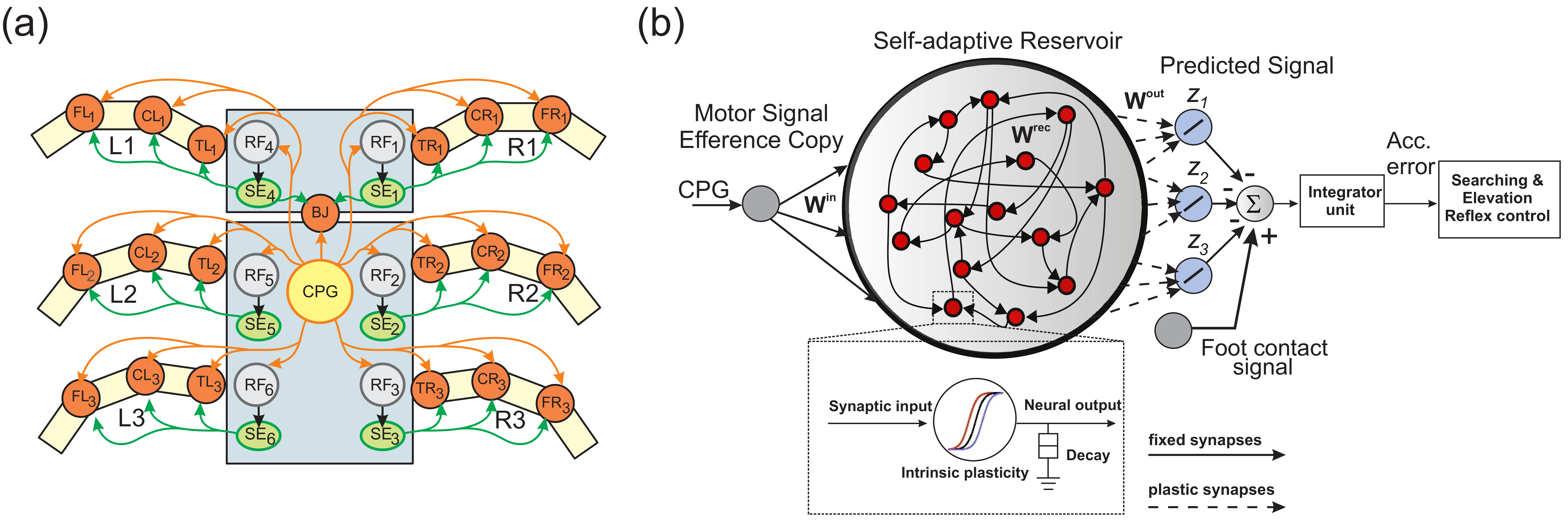} \caption{(a) Neural mechanisms implemented on the bio-inspired hexapod robot AMOSII. The yellow circle ($CPG$) represents the neural locomotion control mechanism (see appendix.~\ref{sec:CPG}). The gray circles ($RF_{1,2,3,...,6}$) represent the reservoir-based adaptive forward models. The green circles ($SE_{1,2,3,...,6}$) represent searching and elevation control modules. The orange circles represent leg joints where $TR_i$, $CR_i$, $FR_i$ are TC-, CTr- and FTi-joints of the right front leg ($i=1$), right middle leg ($i=2$), right hind leg ($i=3$) and $TL_i$, $CL_i$, $FL_i$ are left front leg ($i=1$), left middle leg ($i=2$), left hind leg ($i=3$), respectively. $BJ$ is a backbone joint. The orange arrow lines indicate the motor signals which are converted to joint angles for controlling motor positions. The black arrow lines indicate error signals. The green arrow lines indicate signals for adapting joint movements to deal with different circumstances. b) An example of the reservoir-based adaptive forward model. The dashed frame shows a zoomed in view of a single reservoir neuron. In this setup, the input to each of the reservoir network comes from the CTr-joint of the respective leg. The reservoir learns to produce the expected foot contact signal for three different walking gaits ($z_{1}$, $z_{2}$, $z_{3}$). The signals of the output neurons are combined and compared to the actual foot contact sensory signal. The error from the comparison is transmitted to an integrator unit. The unit accumulates the error over time. The accumulated error is finally used to adapt joint movements through searching and elevation control.}\label{fig:forwardmodels}
\end{center}
\end{figure}

We design, six identical adaptive RNN-based forward models ($RF_{1,2,3,...,6}$), one for each leg of the walking robot (Figure~\ref{fig:forwardmodels}(a)). These serve the purpose of online sensorimotor prediction as well as state estimation. Specifically, each forward model learns to correctly transform the efference copy of the actual motor signal for each leg joint (i.e., here the CTr-joint motor signal\footnote{We use the CTr-joint motor signal instead of the TC- and FTi-motor signals since this shows clear swing (off the ground) and stance (on the ground) phases which can be qualitatively matched to the actual foot contact signal.}), into an expected or predicted sensory signal. This predicted signal is then compared with the actual incoming sensory feedback signals (i.e., here the foot contact signal - Figure~\ref{fig:forwardmodels} (b), of each leg) and, based on the error accumulated over time, it triggers the appropriate action (searching or elevation) and modulate the locomotive behavior of the robot. Each forward model is based on a random RNN architecture of the self-adaptive reservoir network type \citep{dasgupta2013information}, \citep{dasgupta2015thesis}. Due to the presence of rich recurrent feedback connections, the dynamic reservoir and intrinsic homeostatic adaptations, the network exhibits a wide repertoire of nonlinear activity and long fading memory. This can be primarily exploited for the purpose of specific leg joint-motor signal transformation, act as motor memory and for the prediction of sensorimotor patterns arising in the current context.

\subsection*{Network Setup}
The basic setup of each reservoir forward model can be divided into three layers: input, hidden (or internal), and readout layers (Figure~\ref{fig:forwardmodels} (b)). The internal layer consists of a large recurrent neural network driven by time-varying stimuli (CPG motor signals). These driving signals are projected via the input layer. The internal layer is constructed as a random RNN with fixed randomly initialized synaptic connectivity (in this setup we only modify the reservoir-to-readout neuron weights). Using a discrete time version of SARN, with a step size of $\Delta t$, the discrete time state dynamics of each reservoir neuron is given by the following equations: 

\begin{equation}
{x}_i(t+1) = \left(1-\frac{\Delta t}{\tau_i} \right) x_i(t) + \frac{\Delta t}{\tau_i}\left( g\sum_{j=1}^{N}W^{rec}_{i,j}r_j(t) + W^{in}_{i,1}u(t) + B_i\right), i=1,\dots, N. \label{eq:FM_RNN}
\end{equation}

\begin{equation}
 r_i(t) = \mathit{tanh}(a_ix_i(t) + b_i),
\end{equation}

\begin{equation}
 \mathbf{z}(t) = \left[\mathbf{W}^{out}\right]^T \mathbf{r}(t). \label{eq:readout_weight}
\end{equation}

The RNN model consists of $N$ neurons, such that the membrane potential at the soma (at time $t$) of the reservoir neurons, resulting from the incoming excitatory and inhibitory synaptic inputs, is given by a $N$ dimensional vector of neuron state activations. $x(t) = x_1(t), x_2(t),...., x_N(t)$. The RNN here, does not explicitly model action potentials, but describes neuronal firing rates. Where in, the variable $r_i(t)$ describes the instantaneous firing rate ($N$ dimensional) of the reservoir neurons and is calculated as a non-linear function of the state activation $x_i(t)$ (Equation ~\ref{eq:FM_RNN}). Each reservoir neuron $i$, receives inputs from other neurons in the network with firing rates $r_j(t)$ via synaptic connections of strength $W_{ij}^{rec}$ along with incoming stimuli from the input layer via synapses of strength $W_{ij}^{in}$. Each reservoir neuron is also provided with an auxiliary bias $B_i$. The parameter $g$ \citep{sompolinsky1988chaos}, \citep{van1996chaos} acts as the scaling factor for the recurrent connection weights allowing different dynamic regimes from stable ($g<1$) to highly irregular chaotic ($g>1$) \citep{sussillo2009generating}, being present in the network. 

\begin{figure}
\begin{center}
\includegraphics[scale = 0.40]{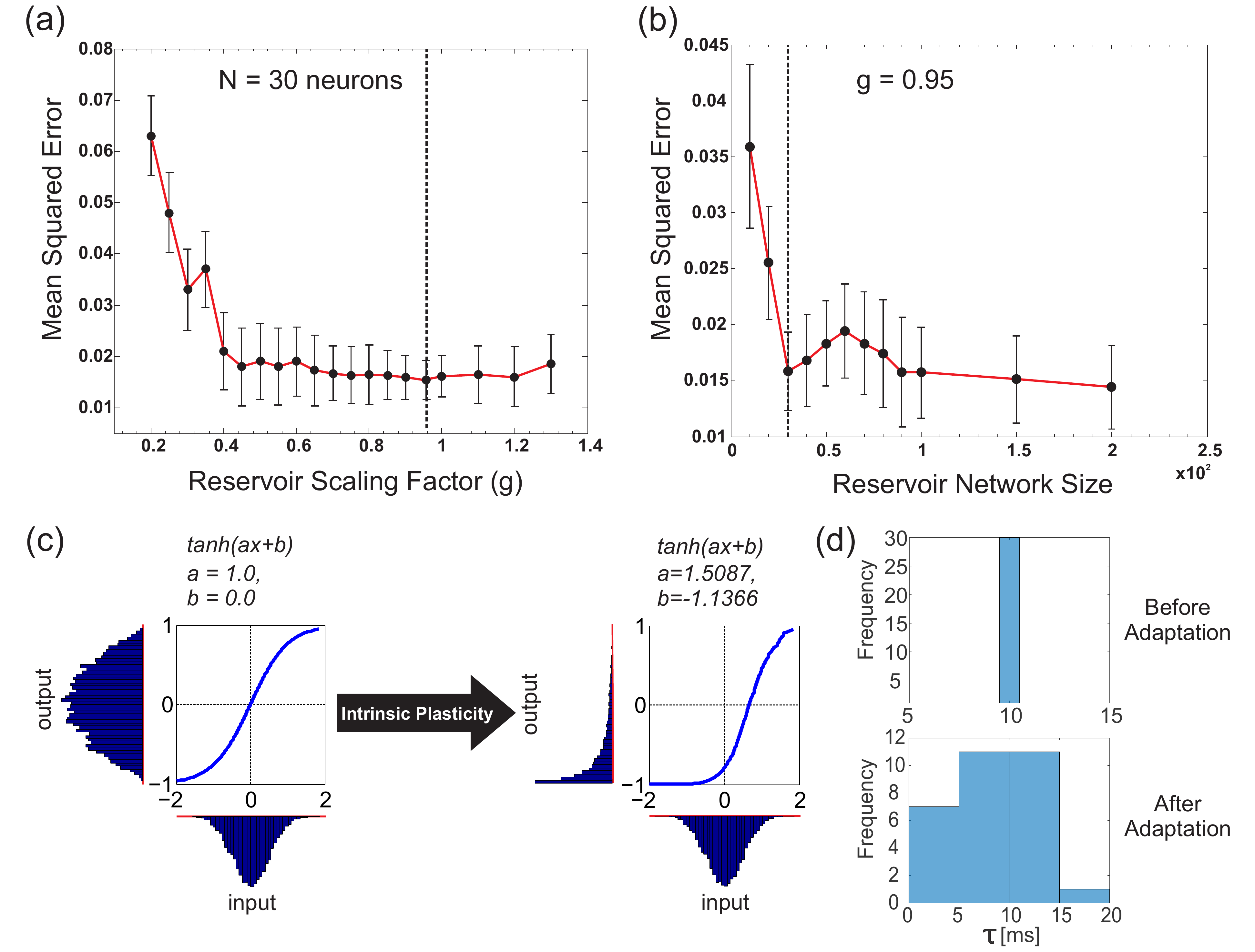} \caption{(a) Plot of the change in the mean squared error for the forward model task for one of the front legs ($R_1$) of the walking robot with respect to the scaling of the recurrent layer synaptic weights $W^{rec}$ with different $g$ values. As observed, very small values in $g$ have a negative impact on performance compared with values closer to one being better. Interestingly, the performance did not change significantly for $g > 1.0$ (chaotic domain). This is mainly due to homeostasis introduced by intrinsic plasticity in the network. The optimal value of $g = 0.95$ selected for our experiments is indicated with a dashed line. (b) Plot of the change in mean squared error with respect to different reservoir sizes ($N$). $g$ was fixed at the optimal value. Although increasing the reservoir size in general tends to increase performance, a smaller size of $N=30$ gave the same level of performance as $N=100$. Accordingly keeping in mind the trade off between network size and learning performance, we set the forward model reservoir size to 30 neurons. Results were averaged over 10 trials with different parameter initializations on the forward model task for a single leg and a fixed walking gait. (c) Example of the intrinsic plasticity to adjust the reservoir neuron non-linearity parameters $a$ and $b$. Initially the the reservoir neuron fires with an output distribution of Gaussian shape matching that of the input distribution. However after adjustment using intrinsic plasticity mechanism \citep{dasgupta2013information} the reservoir neuron adapts the parameters $a$ and $b$, such that, now for the same Gaussian input distribution the output distribution follow a maximal entropy Exponential-like distribution. (d) Distribution of the reservoir forward model individual neuron time constants before and after adaptation.} \label{fig:spectralsize}
\end{center}
\end{figure}  

The input to the reservoir $u(t)$, consists of a single CTr-joint motor signal. This acts as an efference copy of the post-processed CPG motor output. The readout layer consists of three neurons, with their activity being represented by the three-dimensional vector $\mathbf{z}(t)$. Although typically $M <N$ readout neurons can be connected to the reservoir, here we restricted it to three neurons, as each readout here learns the predictive signal for one of the following different walking gaits:  wave ($z_1$), tetrapod ($z_2$), and caterpillar ($z_3$) gaits. The wave, tetrapod, and caterpillar gaits are used for climbing over an obstacle, walking on uneven terrain, and crossing a large gap, respectively\footnote{These three gaits were empirically selected among $19$ other possibilities. Previous studies have demonstrated that the wave and tetrapod gaits are the most effective for climbing and walking on uneven terrains, respectively. While in this particular study we observed that the caterpillar gait was the most effective for crossing a gap. However, without any loss of performance, additional walking gaits can be applied easily by adding further readout neurons.}. Subsequent to the supervised training of the reservoir-to-readout connections $\mathbf{W}^{out}$, each readout neuron basically learns to predict the expected foot contact signal associated with each of these gaits. The decay rate for each reservoir neuron is given by $\frac{1}{\tau_i}$, where $\tau_i$ is the individual membrane timeconstant. The input-to-reservoir connections weights $\mathbf{W}^{in}$ and internal recurrent weights $\mathbf{W}^{rec}$ were drawn randomly from the uniform distribution $[-0.1,0.1]$ and a Gaussian distribution of zero mean and variance $\frac{g^2}{\sqrt{p_cN}}$, respectively. Where, the parameter $p_c$ controls the probability of connections inside the recurrent layer and is set to be 20\%. In order to select the appropriate reservoir size, empirical evaluations were carried out (Figures 3(a) and (b)), such that we achieved a moderate network size of $N=30$, for which the minimum prediction error was obtained at the readout layer, irrespective of the walking gait. The recurrent weights were subsequently scaled by the factor of $g=0.95$ (see Figure~\ref{fig:spectralsize}), such that the spontaneous network dynamics is in a stable regime and achieves the best performance of the chosen network size.  In accordance with the SARN model, unsupervised intrinsic plasticity \citep{triesch2005gradient} and neuron timescale adaptation \citep{dasgupta2015thesis} were carried out in order to learn the transfer function parameters ($a_i$ and $b_i$)and the reservoir timeconstant parameters $\tau_i$ for each individual neuron (Figure~\ref{fig:spectralsize} (c) and (d)). 

\subsection*{Readout Weight Adaptation}

Here we used a modified version of the original recursive least squares (RLS) algorithm \citep{jaeger2004harnessing},\citep{simon2002adaptive} based on the FORCE learning formulation \citep{sussillo2009generating}, in order to learn the reservoir-to-readout connection weights $\mathbf{W}^{out}$ at each time step, while the CPG input ${u}(t)$ is being fed into the reservoir. The readout weights $\mathbf{W}^{out}$ are calculated such that the overall error at the readout neurons is minimized; thereby the network can learn to accurately transform the CTr-motor signal to the expected foot contact signal, for each walking gait. The instantaneous error signal ($e(t)$) at the readout layer, can be calculated as the  difference between the reservoir predicted output ($z(t)$) and the desired output, $d(t)$ (i.e. here the expected foot contact signal). Based on Equation~\ref{eq:readout_weight}, this can be formulated as: 

\begin{equation}
 e(t) = \sum_{j=1}^3 W_j^{out}(t-1)r_j(t) - d(t).
\end{equation}

Using the RLS algorithm, and minimizing this error, the readout weights $W_j^{out}$ update can be defined by,

\begin{equation}
 W_i^{out} = W_i^{out}(t-1) - e(t)\sum_j P_{ij}(t)r_j(t).
\end{equation}

Where, $\mathbf{P}$ is a $N \times N$ square matrix proportional to the inverse of the correlation matrix of the reservoir neuron firing rate vector $\mathbf{r}$. $\mathbf{P}$ is initialized using the identity matrix $\mathbf{I}$ and a small constant parameter $\delta_c$ as, $\mathbf{P} (0) = \frac{\mathbf{I}}{\delta_c}$. $\mathbf{P}$, here, acts as the adaptive learning rate for updating the readout weights with weight modifications automatically slowing down as $\mathbf{P}$ decreases with time. This allows the learning to occur stably and eventually converge to a solution. $\mathbf{P}$ is updated as each time point as, 

\begin{equation}
 \mathbf{P}(t) = \mathbf{P}(t-1) - \left( \frac{\mathbf{P}(t-1)\mathbf{r}(t)\mathbf{r}^T(t)\mathbf{P}(t-1)}{1+\mathbf{r}^T(t)\mathbf{P}(t-1)\mathbf{r}(t)}\right).
\end{equation}

The reservoir-to-readout neuron weights were initialized to zero at start. Details of all the fixed parameters and initial settings for the reservoir based forward model networks are summarized in Supplementary Table 1.


\section{Results}
\subsection{Learning the Reservoir Forward Model (motor prediction)} \label{sec:learn_fm}
In order to train the six forward models ($RF_1 to RF_6$) in an online manner, one for each leg, we let the simulated robot AMOSII walk under normal conditions (i.e., walking on a flat terrain with the three different gaits). Initially, we let the robot walk with a certain walking pattern, and then every $2500$ time steps, the gait pattern was sequentially altered (this occurs by changing the modulatory input to the CPG - see supplementary Figure 1). As a result, the robot sequentially transitions from wave gait, to tetrapod gait, to caterpillar gait repeatedly. Using this procedure, we let the robot walk for three complete cycles ($22500$ time steps) and collected the corresponding CTr-motor signal and foot contact sensor readings for all legs. Intrinsic plasticity and neuron time constant adaptations \citep{dasgupta2013information}, \citep{dasgupta2015thesis}, were then carried out using $20$ epochs of $1000$ time steps overlapping time windows. After this pre-training phase, all the reservoir neuron non-linearity parameters and individual time constants ($\tau_i$) were fixed (see Figure~\ref{fig:spectralsize} (d) for the distribution of neuronal time constants before and after training). 

\begin{figure}
\begin{center} 
\includegraphics[scale = 0.45]{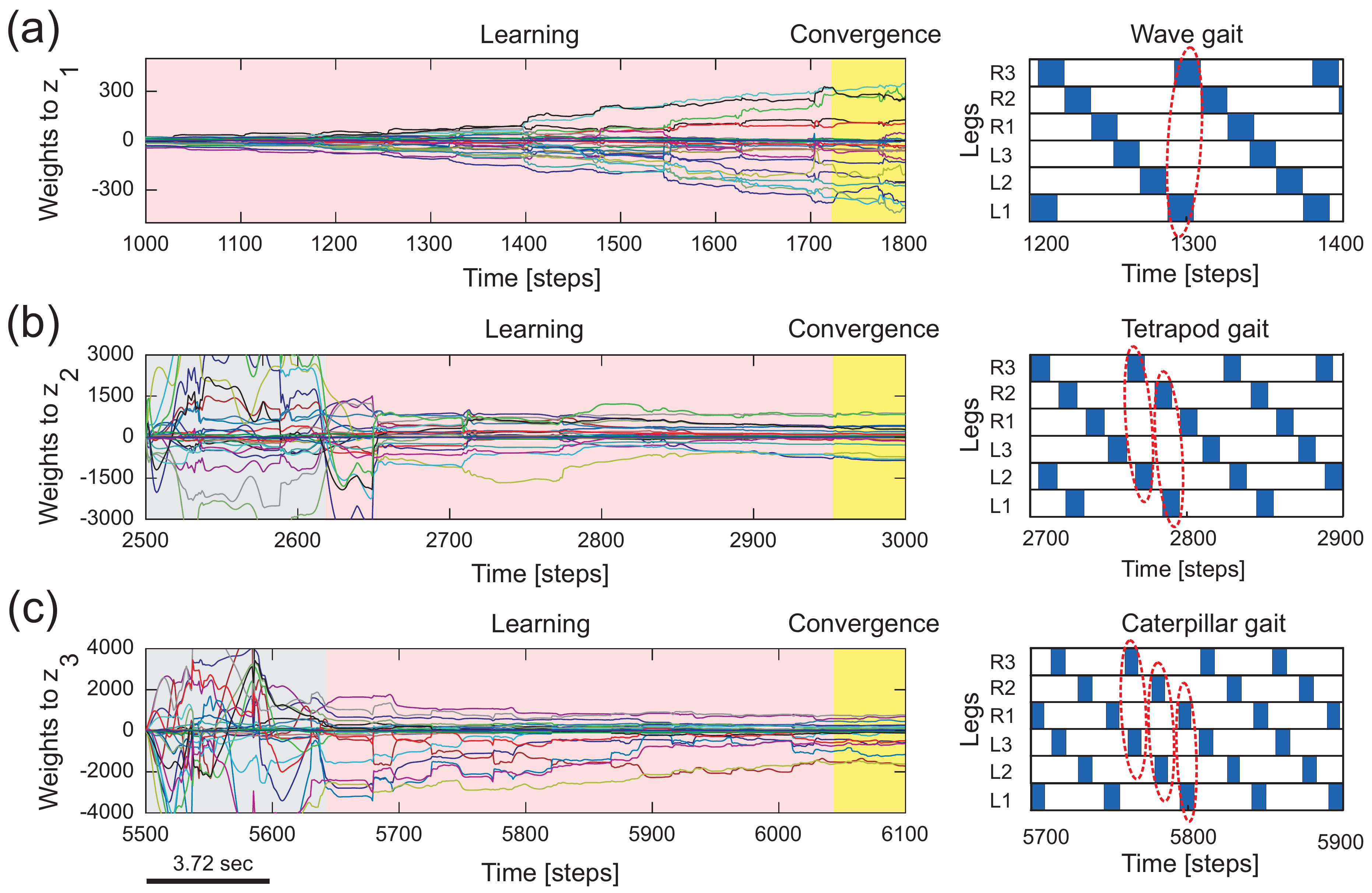} \caption{\textbf{Reservoir-to-readout weight adaptation during online learning.} (a) Changes of 30 weights projecting to the first readout neuron ($z_1$) of the forward model of the right front leg ($R_1$) while walking with a wave gait. During this period, weights projecting to the second ($z_2$) and third ($z_3$) output neurons remain unchanged (i.e., they are zero). (b) Changes of the weights to $z_2$ while walking with a tetrapod gait. During this period, the weights to $z_3$ still remain unchanged and the weights to $z_1$ converge to around zero. (c) Changes of the weights to $z_3$ while walking with a caterpillar gait. During this period, the weights to $z_1$ and $z_2$ converge to around zero. At the end of each gait, all weights are stored such that they will be used for locomotion in different environments. The grey areas represent transition phases from one gait to another gait and the yellow areas represent convergence. The gait diagrams are shown on the right. They are observed from the motor signals of the CTr-joints (Figure~\ref{fig:signaltransformation}). White areas indicate ground contact or stance phase and grey areas refer to no ground contact during swing phase. As frequency increases, some legs step in pairs (dashed enclosures). Here convergence implies no siginificant change in the vector norm of the readout weights.}\label{fig:fm_weights}
\end{center}
\end{figure}

Subsequent to the pre-training phase, normal training of the reservoir-to-readout weights $\mathbf{W}^{out}$ was carried out using the online RLS learning algorithm with the same process of making the robot walk on a flat, regular terrain and sequential switching between the three gait patterns every $2500$ time steps. As such, at any given point in time only one of the readout neurons (specific to the walking gait) are active. In this manner, synaptic weights projecting from reservoir to the first readout  neuron ($z_{1}$) corresponding to the foot contact signal prediction for the wave gait, and synaptic weights projecting to the second ($z_{2}$) and third ($z_{3}$) readout neurons corresponding to the foot contact signal prediction of the tetrapod and caterpillar gaits, are learned, respectively. Within this experimental setup, as observed from Figures~\ref{fig:fm_weights} (a), (b) and (c) the readout weights corresponding to each gait converges very quickly (due to intrinsic noise and nature of the reservoir-to-readout synaptic adaptation, the weights still show minute fluctuations after successful learning; therefore here convergence applies that the norm of the readout weights $|W^{out}|$ remains constant with a small finite value \citep{sussillo2009generating}), in less than the trial period of $2500$ time steps. As a result, every time the CTr-motor signal changes due to walking gait transformations, the RF associated with each leg learns to predict the expected foot contact signal robustly. The training process was carried out only once under normal walking conditions. This was subsequently used as the baseline in order to compare with the actual foot contact signals (sensory feedback) while walking under the situations of crossing a gap, climbing, and negotiating uneven terrains. 

\begin{figure}
\begin{center} 
\includegraphics[scale = 0.5]{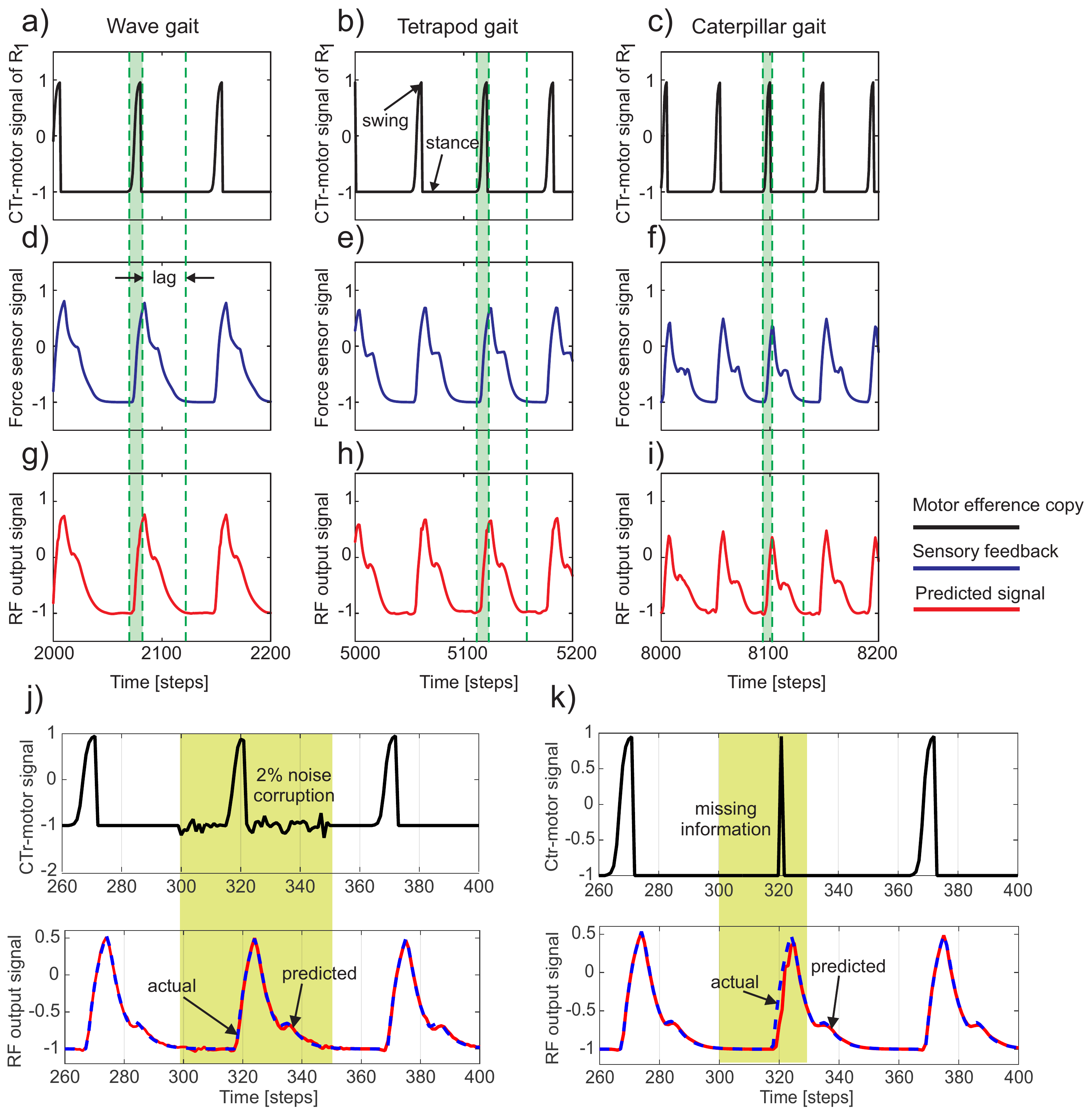} \caption{(a-c) The CTr-joint motor signal of the right front leg ($R_1$) for wave, tetrapod, and caterpillar gaits, respectively. This motor signal provides the efference copy or the input to the reservoir forward models. (d-f) The actual foot contact signal (force sensor signal under normal walking conditions) used as the target signal of the reservoir models. (g-i) The predicted foot contact signal or the final learned output of the forward model for each walking gait ($RF$ output signal). The green shaded region indicates the time interval between swing and stance phase for the CTr motor signal at the three walking gaits. As observed the actual foot contact signal is considerably lagged in time compared to the motor signal. Effectively, this lag decreases with an increase in the gait frequency. The single RF adaptively accounts for these different delay times in order to accurately predict the expected foot contact signal. (j) above - CTr-joint motor signal demonstrated for a single leg, with 2\% Gaussian noise injected between 300-350 time steps (yellow shaded region), below - Despite the noise corruption of the motor signal, the reservoir forward model is able to generate the correct predicted FC signal (blue dotted - target FC signal, red solid - predicted signal). (k) above - The CTr-joint motor signal corrupted with missing information between 280-320 time steps. As a result, the motor signal shows a narrow spike between 310 -330 time steps (yellow shaded region), below - Reservoir forward model predicted signal (red) as compared to the desired FC signal (dotted blue). Although the CTr motor signal was transiently missing, the reservoir is able to generate the desired FC signal considerably well, while at the same time maintaining the correct temporal sequence of the signals. }\label{fig:signaltransformation}
\end{center}
\end{figure}

Figure~\ref{fig:signaltransformation} shows an example of the forward model prediction (training) during the three different walking gaits, for the right front leg of AMOSII ($R_1$). Visual inspection clearly demonstrates that according to the corresponding efference copy of CTr-motor signal at a particular gait, the expected foot contact (FC) signal is precisely predicted at each time point. Similarly, the foot contact signals for the other legs are also predicted online, given the current context of CTr-signal (not shown). Note that the FC signals of the other legs normally show slightly different periodic patterns. Furthermore, there exists considerable lag between the expected stance phase according to the motor signal and that observed from the FC signal (difference between dotted green lines in Figure ~\ref{fig:signaltransformation}). Due to the internal memory of the incoming motor signal in the reservoir, we see that the output neurons can adapt to these time lags efficiently, even when the frequency of the signal increases with a change in walking gaits. Furthermore, the reservoir-based forward models enable the robust generation of the predicted FC signal, even in the presence of high noise corruption or missing information in the incoming CTr-joint motor signal (Figures~\ref{fig:signaltransformation} (j) and (k)). Due to the fact that the CTr-motor signals are obtained after appropriate post-processing of original CPG singals and passage through the motor neurons coupled with different time delays. Such signal corruption can occur at various levels. Therefore, the ability of the forward model to deal with such abrupt noise in the motor signals in a robust manner is crucial to the adaptive mechanisms. Furthermore, such signal corruptions can also occur, due to entrainment mechanisms applied for the automatic tuning or adaptation of CPG outputs \citep{nachstedt2013adaptive}. Such online adaptation for sudden motor signal variations, was not possible in the previous state of the art adaptive neuron forward models \citep{manoonpong2013neural}. This model inherently lacked the ability to deal with variations in the temporal properties of the signal. As such, a simple square wave matching the timing of the motor signal efference copy was used, providing a limited range of behavior, as well as being biologically implausible. However, here our reservoir-based model can accurately estimate the spatiotemporal properties of the signal and robustly learn the exact shape, as well as the timing of the actual FC signals.  

\subsection{Simulated Complex Environments} \label{sec:fm_expresults}
In order to assess the ability of the reservoir-based forward models to generate adaptive\footnote{Forward models for motor prediction need an internal fading memory of the motor apparatus, in order to adjust for time delays between motor output signal and the actual sensory feedback \citep{kawato1999internal}.} complex locomotive behaviors in a neural closed-loop control system (see Figure~\ref{fig:neuralmechanisms}), we conducted simulation experiments under different situations including crossing a gap, walking on uneven terrain and climbing over high obstacles (similar to the behaviors observed in real insects). In all cases, we used the same training procedure for the forward models by allowing the robot to walk under normal conditions on a flat even terrain. 

\begin{figure}
\begin{center} 
\includegraphics[width=\linewidth]{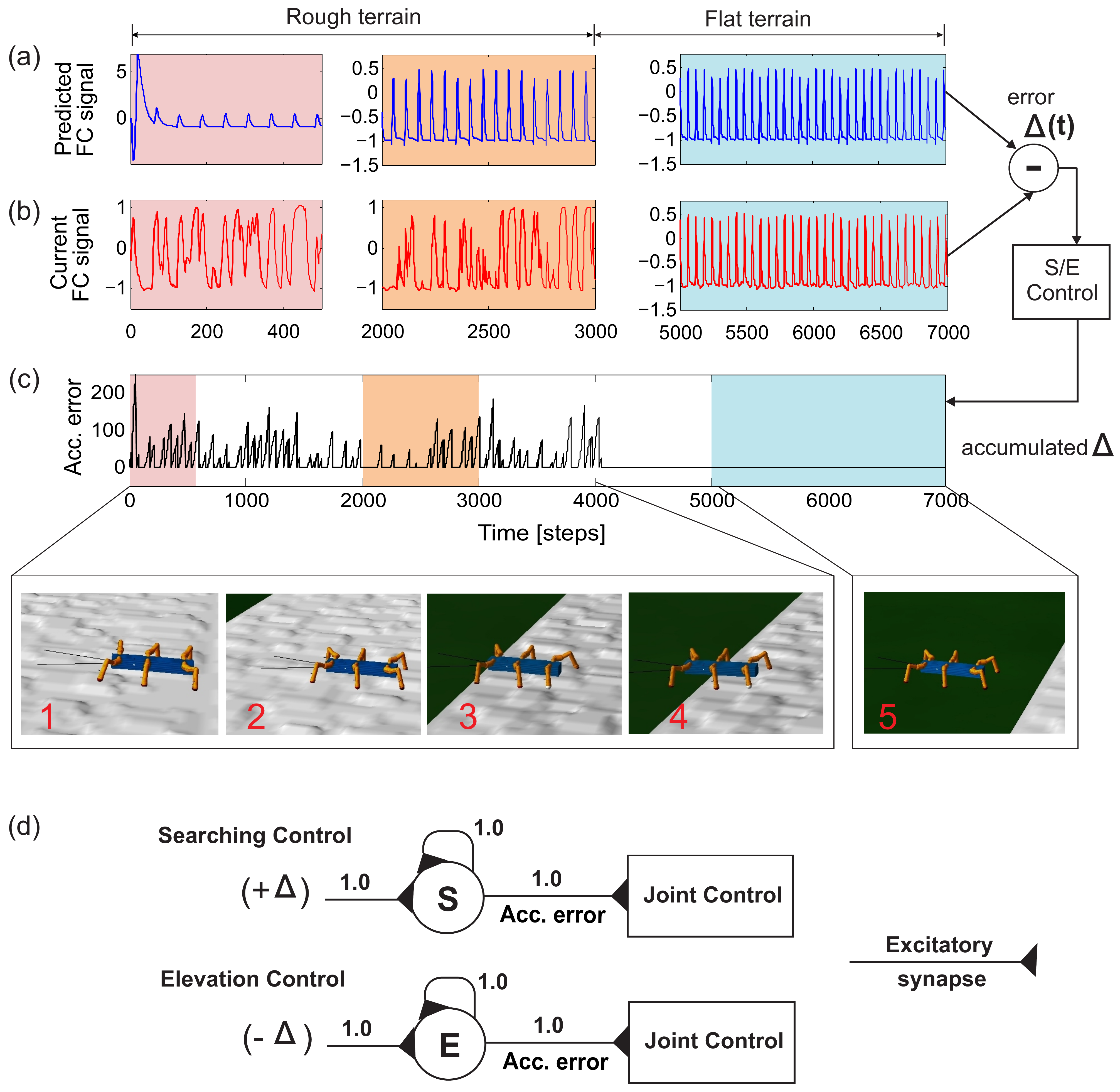} \caption{\textbf{Successfully navigating rough terrain with reservoir forward model} (a) The reservoir forward model predicted, expected foot contact signal. After a small initial transient the reservoir output quickly converges to the expect signal for normal walking condition. (b) The actual sensory feedback (foot contact signal) while walking on the rough surface (c) Accumulated error calculated from the instantaneous error ($\Delta(t)$) after passing through the recurrent neuron in the searching and elevation control . (d) The searching and elevation action control system consisting of individual recurrent neurons as signal accumulators. After 4000 time steps, the robot successfully overcomes the rough terrain and continuous walking on a flat surface. As a result, there is zero accumulated error since the predicted foot contact signal almost exactly matches the actual signal. See the experiment supplementary video 3.}\label{fig:testing_FM}
\end{center}
\end{figure}

During testing of the learned behavior, while AMOSII walks under different environmental conditions and a specific gait, the output of each trained forward model (i.e., the predicted FC signal, Figure~\ref{fig:testing_FM} (a)) is used to compare it to the actual incoming FC signal of the leg (Figure~\ref{fig:testing_FM} (b)). The difference (instantaneous error signal $\Delta$) between them determines the walking state where a positive value ($+\Delta$) indicates losing ground contact during the stance phase and a negative value ($-\Delta$) indicates stepping on or hitting obstacles during the swing phase. 
\begin{equation}
\Delta_i(t) = RF_i(t)-FC_i(t). \label{eq:inst_error}
\end{equation}
where $i \in \{1,2,...,6\}$ represents each leg of the robot. 

\begin{figure}
\begin{center} 
\includegraphics[scale=0.5, width=\linewidth]{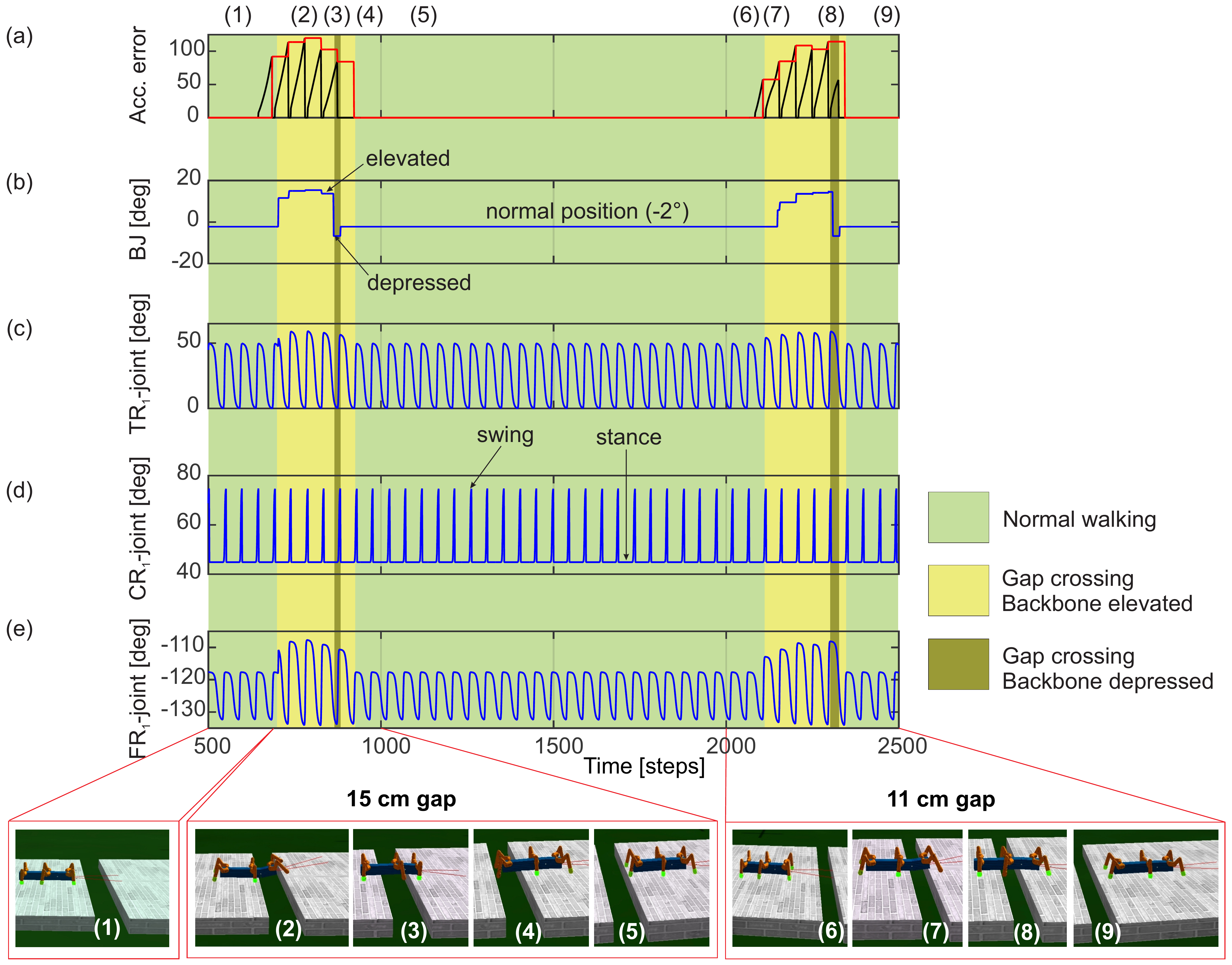} \caption{\textbf{Real-time data of walking and crossing multiple gaps using the forward model predictions.} (a) The accumulated error (black line) and the maximum accumulated error value at the end of each stance phase (red line) of the right front leg ($R_1$). The accumulated error is reset to zero every swing phase. (b) The backbone joint (BJ) angle during walking and gap crossing. The BJ stays at the normal position ($-2 \deg$) during normal walking. On encountering a gap ($15$cm), it leans upwards in a step like fashion and then finally bent downwards in order to cross the gap. This procedure is repeated for the second gap ($11$cm), however with different degree of elevations. (c-e) The TC-, CTr-, and FTi-joint angles of right front leg $R_1$ during normal walking and gap crossing. The joint adaptation was controlled by the maximum accumulated error value of the  previous step (red line). Below pictures show snap shots of the locomotion of AMOS II during the experiment. Note that one time step is $\approx 0.037$~s. For further details interested readers are recommended to see the experiment supplementary videos 1 and 2.}\label{fig:Gap_crossing}
\end{center}
\end{figure}

Thus, we use the positive value for searching control (Figure~\ref{fig:testing_FM} (d) above). This is then accumulated through a single recurrent neuron $S$ with a linear transfer function and is always reset to 0.0 at the beginning of swing phase. Similarly, the negative value is used for elevation control (Figure~\ref{fig:testing_FM} (d) below). The value is also accumulated through a recurrent neuron $E$ with a linear transfer function. These accumulated errors (Figure~\ref{fig:testing_FM} (c)) thus allow the robot leg to be either elevated (on hitting an obstacle) or searching for a foothold during the swing and stance phases respectively (see \citep{manoonpong2013neural} for more details of the searching and elevation control). As depicted in Figures~\ref{fig:testing_FM} (a) and (b), while walking on a rough terrain (in this case with tetrapod walking gait), the currently recorded sensory feedback or foot contact sensor reading differs considerably from the reservoir predicted signal. As a result, there is a high accumulation of error between each swing or stance phase (Figure~\ref{fig:testing_FM} (c)). It should be noted that the initial ($\approx 50$ time steps) abruptly high amplitude signal observed in the reservoir forward model prediction, is caused due to the transient recovery time needed by reservoir readout neurons to settle to the exact learned patterns. This is overcome within the next few time steps and RF predicted FC signal continues to occur in a robust manner. The accumulated error causes the corresponding leg action control mechanism to kick in and the robot successfully navigates out of the rough terrain (after $\approx 4000$ time steps). Once the robot moves into the flat terrain, the reservoir predicted foot contact signal matches almost perfectly with the actual sensory feedback. As a result, the accumulated error becomes zero and normal walking without any additional searching or elevation control mechanisms, can continue. In essence based on the reservoir forward models, while traversing from the uneven terrain (Figure~\ref{fig:testing_FM} inset 1-4) to the flat terrain (Figure~\ref{fig:testing_FM} inset 5), the robot can adapt its legs individually to deal with the change of terrain. That is, it depressed its leg and extended its tibia to search for a foothold when loosing ground contact during the stance phase. Losing ground contact information is detected by a significant change of the accumulated errors (Figure~\ref{fig:testing_FM} (c)). In case of both walking on uneven terrain and climbing, this accumulated error causes shifting of the CTr- and FTi-joints causing the respective leg to search for a foothold. However, in the specific case of crossing a gap (Figure~\ref{fig:Gap_crossing}), we use the accumulated error in order to control tilting of the backbone joint (BJ) and shifting of the TC- and FTi-joints such that the front legs can be extended forward continuously till the robot can find a foothold. In addition to this leg joint control, reactive backbone joint control using the additional ultrasonic sensors in front of the robot can also be used to learn to lean up the BJ for climbing over obstacles (this has been previously successfully applied using classical conditioning based learning in \citep{goldschmidt2014biologically} and as such not discussed here). 

We now take the example of the more complex, multiple gap crossing experiment in order to look in detail at the learning outcome of the forward models. This experiment was divided into two components, consisting of one larger gap ($15$cm length) and another relatively shorter gap of $11$ cm length. The two gaps were separated by considerable distance where the robot was allowed to walk on a regular flat terrain. In order to learn to cross a gap, we let AMOS II walk with a caterpillar gait (see Figure~\ref{fig:fm_weights} (c), right), such that each left and right pair of legs moves simultaneously. Empirically this is observed to be the most suited gait for overcoming large gaps, as well as supported by experimental observations in stick insects ~\citep{blaesing2004stick}. As shown in Figure~\ref{fig:Gap_crossing}(1), at the beginning AMOS II walked forward straight towards the initial gap. In this period, as it walks on the flat surface of the platform, it performed regular movements similar to the training period under normal walking conditions (training on a flat regular surface) . Eventually, it encounters a $15$ cm wide gap ($\approx$ 44$\%$ of body length - the maximum cross-able distance). In this situation, during the subsequent stance phase the front legs of the robot loose ground contact (Figs.~\ref{fig:Gap_crossing}(d) and (e)). As a result, the foot contact sensors from the front legs do not record any value. However the reservoir forward model still predicts the expected foot contact signal, causing a positive instantaneous error (Eq.~\ref{eq:inst_error}). This leads to a gradual ramping of the accumulated error signal between each stance phase and swing phase, for the front legs (Figure~\ref{fig:Gap_crossing} (a)).

In order to activate the BJ and adapt the leg movements due to the  difference between the reservoir predicted FC signal and the actual sensory feedback of the FC sensors (error signals), we used the maximum accumulated error value of the previous step (Figure~\ref{fig:Gap_crossing}, (a) red line) and control the BJ and leg movements in the subsequent step. In this manner, the BJ started to lean upwards incrementally (step like manner) at around $680-850$ time steps (Figure~\ref{fig:Gap_crossing}(2)). Simultaneously, the TC- and FTi-joint movements of the left and right front legs were also adapted accordingly in order to carry out elevation action (this is reflected in the higher amplitude of these two signals in this time period). Due to a predefined time-out period for tilting upwards, at around $850$ time steps (Figure~\ref{fig:Gap_crossing}(3)), the backbone joint automatically moved downwards recording a negative value. Consequently, the front legs touch the ground of the second platform at the middle of the stance phase; thereby, causing the accumulated error signals to decrease. Due to another time-out period for tilting downwards at around $900$ time steps (Figure~\ref{fig:Gap_crossing}(4)), the BJ automatically moved to the normal position ($-2 \deg$). Since now the situation is similar to walking on flat terrain, the RF predicted foot contact signal matches the one recorded by the foot sensors, with accumulated error dropping to zero. Thereafter, the TC- and FTi-joints perform regular movements. Subsequently left and right hind legs loose the ground contact, and AMOSII continues to walk forward. Here the movements of the TC- and FTi-joints were slightly adapted allowing AMOS II to successfully cross the gap and continue walking on the second platform (Figure~\ref{fig:Gap_crossing}(5)). As the terrian now resembles a regular flat surface (similar to the original training terrain) AMOSII two continues to walk forward in normal manner  with no accumulated errors being present. However, the same procedure is repeated once again, when AMOSII re-encounters the second gap at around $2100$ time steps. However in this case, since the gap length is much smaller, the elevation in the BJ occurs with an initial increment of smaller amplitude (Figure~\ref{fig:Gap_crossing} (2)) as compared to the previous case. Thereafter, a similar process is followed and AMOSII can once again successfully overcome this gap and continue walking on the other end of the platform (Figure~\ref{fig:Gap_crossing} (9)). This clearly demonstrates the adaptive yet robust performance of the forward model based predictions in order to successively cross gaps of different length.

\begin{figure}
\begin{center} 
\includegraphics[scale = 0.28]{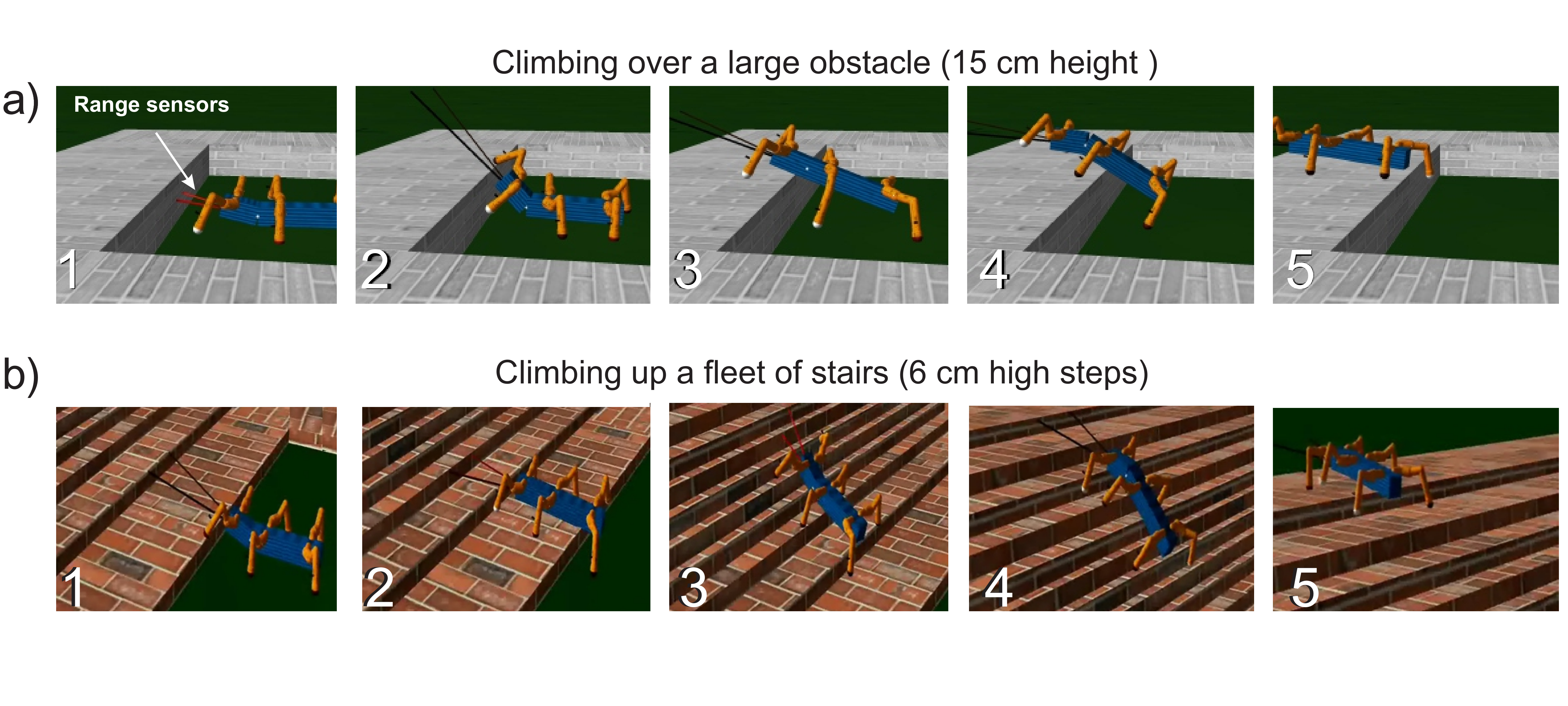} \caption{\textbf{Snapshots showing the learned behavior during climbing over a high obstacle and climbing up a fleet of stairs} (a) AMOSII walked with the wave gait and approached a 15 cm high obstacle (1). It detected the obstacle using its range sensors installed at its front part. The low-pass filtered range sensory signals control the BJ to tilt upwards (2) and then back to its normal position (3). Due to the missing foot contact of the front legs, the BJ moved downwards to ensure stability (4). During climbing, middle and hind legs lowered downwards due to the occurrence of the accumulated errors, showing leg extension, to support the body. Finally, it successfully surmounted the high obstacle (5). For further details see the supplementary experiment video 4 (b) AMOSII climbed up a fleet of stairs (1-5) using the wave gait as well as the reactive BJ control. The climbing behavior is also similar to the one described in the case (a). For further details see supplementary experiment video 5. }\label{fig:Experiment23}
\end{center} 
\end{figure}

Figure~\ref{fig:Experiment23} shows that the reservoir forward model in combination with the neural locomotion control mechanisms, not only successfully generates gap crossing behavior of AMOS II (as shown above), and learns to walk on uneven terrain, but also allows it to climb over single and multiple obstacles (eg. up a fleet of stairs).  In all these cases, we directly used the accumulated errors for movement adaptation via the searching and elevation control mechanisms. For climbing, the reactive backbone joint control was also applied to the system (see \citep{goldschmidt2014biologically} for more details) and a slow wave gait walking pattern (see Figure~\ref{fig:fm_weights} (a), right) was used.
 
Experimentally the wave gait was found to be the most effective for climbing, which allows AMOSII to overcome the highest climbable obstacle (i.e., 15 cm height which equals $\approx 86 \%$ of its leg length) and to surmount a fleet of stairs.  For walking on uneven terrain, a tetrapod gait (see Figure~\ref{fig:fm_weights} (b), right) was used without the backbone joint control. This is the most effective gait for walking on uneven terrain (see also \citep{manoonpong2013neural}). Recall that in all experiments the forward models basically generate the expected foot contact signals (i.e., sensory prediction), which are compared to the actual incoming ones. Errors between the expected and actual signals during locomotion serve as state estimation and are used to adapt the joint movements accordingly. It is important to note that, the best gait for each specific scenario was experimentally determined and fixed. However, this could be easily extended with learning mechanisms (see \citep{steingrube2010self}) to switch to the desired gait when the respective behavioral scenarios are encountered, without any additional influence on the performance of the reservoir forward models.

\begin{figure}
\begin{center} 
\includegraphics[width = \linewidth]{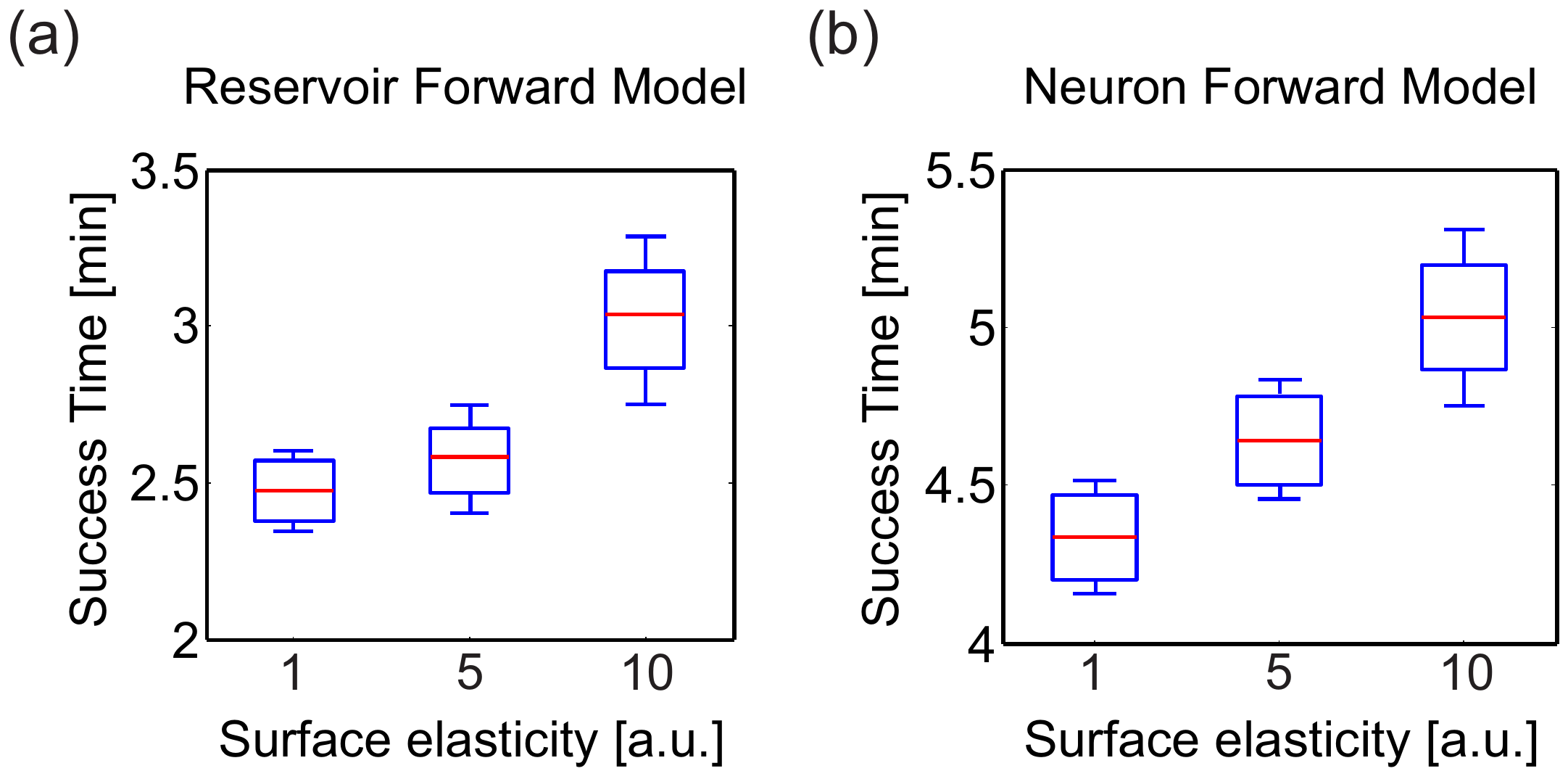} \caption{\textbf{Average time to successfully overcome uneven terrains of different elasticity (hard, moderate, highly elastic)} (a) Average success time for reservoir-based forward model. (b) Average success time for adaptive neuron forward model from \citep{manoonpong2013neural}. Here the whiskers indicate one standard deviation above and below the mean value. Note the difference in scale of the y-axis in both plots. The experimental surface here consisted of the rough terrain as presented in Fig.~\ref{fig:testing_FM} consisting of irregular undulations, however with varying degree of elasticity for the three cases.}\label{fig:compare_elastic}
\end{center} 
\end{figure}

In order to evaluate the performance of our adaptive reservoir forward model in comparison to the state of the art model recently presented in \citep{manoonpong2013neural} (single recurrent neural with low-pass filter), we carried out simulation experiments with AMOSII walking on different types of surfaces. Specifically, after training on a flat surface (under normal conditions) we carried out 10 trials each with the robot walking on uneven terrains (laid with multiple obstacles of height $8cm$), having three different elastic properties\footnote{Here the elasticity coefficients do not strictly represent Young's modulus values. These were local parameter setting defined in the simulation, with increasing values causing greater elasticity.}. The surfaces were divided into hard ($1.0$),  moderately elastic ($5.0$) and highly elastic ($10.0$). A tetrapod walking gait was used in all three cases. Starting from a fixed position, we noted the total time taken by the robot to successfully cross the uneven terrain region and move into a flat surface region. As observed in Figs.~\ref{fig:compare_elastic} (a) and (b), the reservoir forward model enables the robot to traverse the uneven region considerably faster as compared to the adaptive neuron forward model, in all three scenarios. Both the models can be seen to overcome the hard surface much better as compared to the elastic ones. This was expected due to the changes in surface stiffness resulting in additional forces on the robot legs. However, the reservoir model performance was considerably more robust with a mean difference in success time of $1.86$ mins for the hardest surface and approximately $2$ mins for the most elastic surface, cases. Given that the walking gait was fixed, here the success time can be thought as an indicator of the robot's energy efficiency. In the absence of additional body mechanisms to deal with changing surface stiffness, the reservoir based model outperforms the previous implementations of adaptive forward models by $\approx 25\%$ order of magnitude on average.

\section{Discussion} \label{sec:diss4}
In this study, we presented adaptive forward models using the self-adaptive reservoir network for locomotion control. The model is implemented on each leg of a simulated bio-inspired hexapod robot. It is trained online during walking on a flat terrain in order to transform an efference copy (motor signal) into an expected foot contact signal (i.e., sensory prediction). Afterwards, the learned model of each leg is used to estimate walking states by comparing the expected foot contact signal with the actual incoming one. The difference between the expected and actual foot contact signals is used to adapt the robot's leg through elevation and searching control. Each leg is adapted independently. This enables the robot to successfully walk on uneven terrains. Moreover, using a backbone joint, the robot can also successfully cross a large gap and climb over a high obstacle as well as up a fleet of stairs. In this approach, basic walking patterns are generated by CPG-based control along with local leg control mechanisms that make use of the reservoir prediction to adapt the robot's behavior. The key neural mechanisms presented in this work, namely, CPG -based neural control, internal forward models and local leg control, are essential for robust, adaptive locomotion control.  However, only individual instances of them has been successfully realized on artificial and bio-mimetic robotic systems \citep{blasing2004adaptive}, \citep{lewinger2011neurobiologically}, \citep{schilling2012grounding}, \citep{ren2012multiple}, \citep{christensen2014fault}, \citep{pfeifer2007self}; thereby achieving partial solutions. Furthermore, although a few studies have focused on a combination of these neural mechanisms, they have largely been tailored for adaptive locomotion in quadruped robots \citep{lewis2002gait}, \citep{silva2012adaptive}, without the ability to climb obstacles or cross large gaps, as observed in real animals and insects. Thus, this work demonstrates how the combination of these essential components, coupled with the power of the adaptive recurrent neural forward models can achieve very rich behavioral repertoire in bio-inspired hexapod robots. Thus supporting the idea that such embodied neural control \citep{floreano2014robotics} is indeed a potential powerful future alternative of more conventional control methods. 

It is important to note that the usage of reservoir networks, as forward models here, provides the crucial benefit of an inherent representation of time and fading memory (due to the internal feedback loops and input dependent adaptations). Such memory of the time-varying motor or sensory stimuli is required to overcome intrinsic time lags between expected sensory signals and motor outputs \citep{wolpert1998internal}, as well as in behavioral scenarios with considerable dependence on the history of motor output \citep{lonini2009internal}. This is very difficult in most of the previous implementations of forward internal models using either simple single recurrent neuron implementations \citep{manoonpong2013neural}, feed-forward multi-layered neural networks \citep{schroder2010using}, or Bayesian network models \citep{dearden2005learning}, \citep{sturm2008adaptive}.
Furthermore, in this case, online adaptation of only the reservoir-to-readout weights (readout) makes such networks beneficial for simple and online learning. 

The concept of forward models with efference copies in conjunction with neural control has been suggested since the mid-20th century \citep{holst1950reafferenzprinzip}, \citep{held1961exposure} and increasingly employed for biological investigations \citep{webb2004neural}. This is because it can explain mechanisms which biological systems use to predict the consequence of their action based on sensory information, resulting in adaptive and robust behaviors in a closed-loop scenario. This concept also forms a major motivation for robots inspired by biological systems. Within this context, the work presented here, verifies that a combination of CPG-based neural control, adaptive reservoir forward models with efference copies, and searching and elevation control can be used for robustly generating complex locomotion and adaptive behaviors in an artificial walking system. Additionally, although in this study we specifically focused on locomotive behaviors for walking robots, (such) SARN based motor prediction systems can be easily generalized to a number of other applications. Specifically for neuro-prosthetic \citep{ganguly2009emergence}, sensor-driven orthotic control \citep{braunwoergoettermanoonpong2014a}, \citep{lee2005gait} or brain-machine interface devices \citep{golub2012internal}, that require the learning of such predictive models using highly non-stationary, temporal signals, applying SARN models can provide high performance gains with embedded memory, as compared to the current static feed-forward neural network solutions. In the future, we will transfer the reservoir-based adaptive forward models to the physical hexapod robot AMOS-II \citep{manoonpong2013neural} in order to test the adaptive behaviors in a real environment. Furthermore, although in this work, we specifically focused on a single CPG-based control mechanism, in the future we plan to augment the distributed forward model architecture with multiple CPG-based control (one for each leg) \citep{ren2015multiple}. Thereby, truly enabling decentralized control of the robot legs for greater degree of adaptation as observed in biology. 

\section*{Acknowledgement}
This research was supported by the Emmy Noether Program (DFG, MA4464/3-1), the Federal Ministry
of Education and Research (BMBF) by a grant to the Bernstein Center for Computational Neuroscience II G\"{o}ttingen (01GQ1005A, project D1) and the International Max Planck Research School for Physics of Biological and Complex Systems scholarship.

\section*{Author contributions}
S.D, F.W and P.M designed the research. S.D and P.M implemented the model, analyzed data and carried out simulations. D.G carried out the climbing experiments. S.D and P.M wrote the manuscript.

\bibliography{References}

\includepdf[pages=-, pagecommand={}]{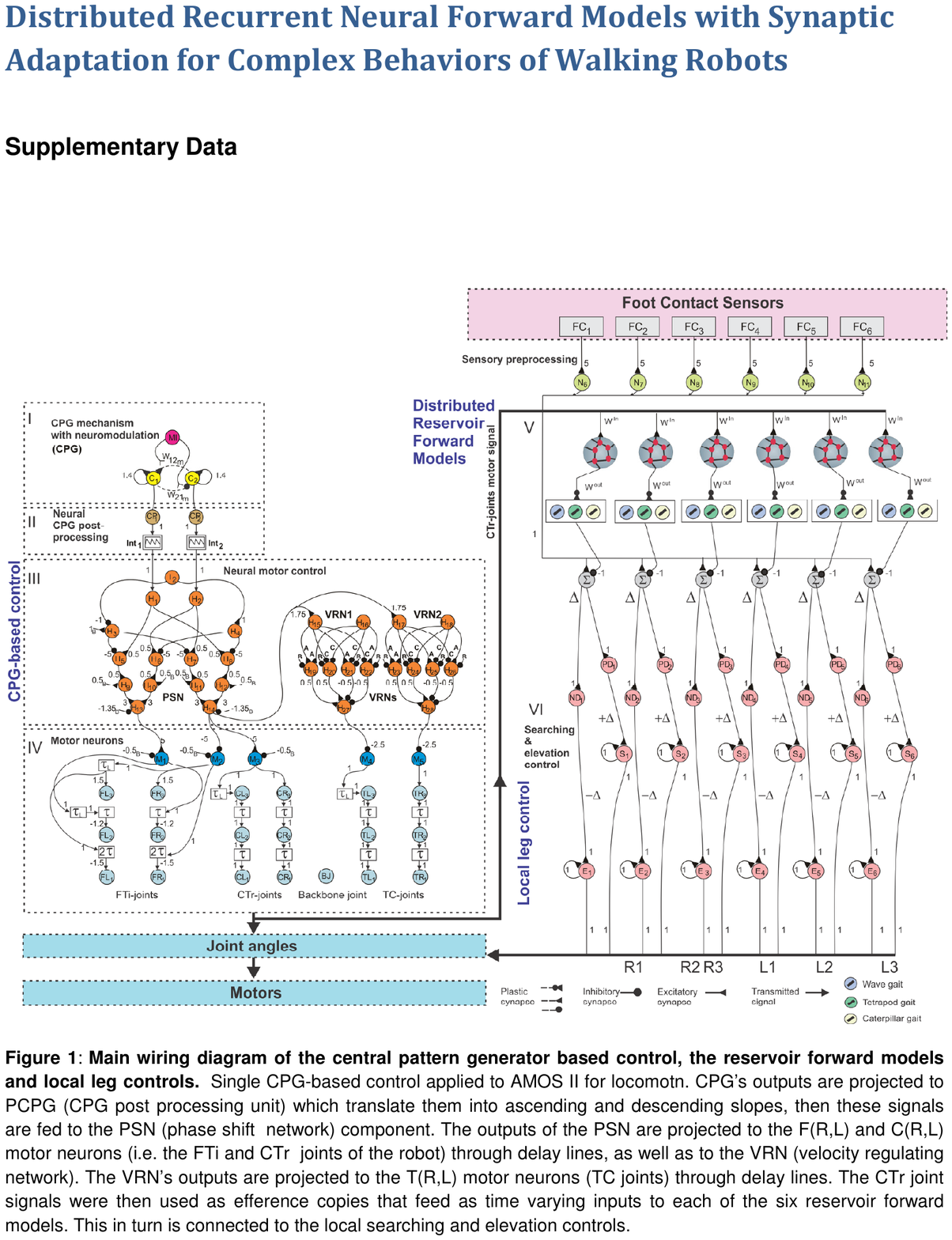}

\end{document}